\documentclass[sigconf]{acmart}
\usepackage{array}
\usepackage{CJKutf8}
\usepackage{subfigure}
\usepackage{multirow}

\AtBeginDocument{%
  \providecommand\BibTeX{{%
    \normalfont B\kern-0.5em{\scshape i\kern-0.25em b}\kern-0.8em\TeX}}}

\copyrightyear{2019}
\acmYear{2019}
\setcopyright{acmcopyright}
\acmConference[MMAsia '19]{ACM Multimedia Asia}{December 15--18, 2019}{Beijing, China}
\acmBooktitle{ACM Multimedia Asia (MMAsia '19), December 15--18, 2019, Beijing, China}
\acmPrice{15.00}
\acmDOI{10.1145/3338533.3366579}
\acmISBN{978-1-4503-6841-4/19/12}

\acmSubmissionID{56}

\begin{document}
\title{A Cascade Sequence-to-Sequence Model for Chinese Mandarin Lip Reading}

\author{Ya Zhao}
\affiliation{
 \institution{Zhejiang Provincial Key Laboratory of Service Robots}
 \institution{Zhejiang University}
}
\email{yazhao@zju.edu.cn}

\author{Rui Xu}
\affiliation{
 \institution{Zhejiang Provincial Key Laboratory of Service Robots}
 \institution{Zhejiang University}
}
\email{ruixu@zju.edu.cn}

\author{Mingli Song}
\affiliation{
 \institution{Zhejiang Provincial Key Laboratory of Service Robots}
 \institution{Zhejiang University}
}
\email{brooksong@zju.edu.cn}

\begin{abstract}
Lip reading aims at decoding texts from the movement of a speaker's mouth. In recent years, lip reading methods have made great progress for English, at both word-level and sentence-level. Unlike English, however, Chinese Mandarin is a tone-based language and relies on pitches to distinguish lexical or grammatical meaning, which significantly increases the ambiguity for the lip reading task. In this paper, we propose a Cascade Sequence-to-Sequence Model for Chinese Mandarin (CSSMCM) lip reading, which explicitly models tones when predicting sentence. Tones are modeled based on visual information and syntactic structure, and are used to predict sentence along with visual information and syntactic structure. In order to evaluate CSSMCM, a dataset called CMLR (Chinese Mandarin Lip Reading) is collected and released, consisting of over 100,000 natural sentences from China Network Television website. When trained on CMLR dataset, the proposed CSSMCM surpasses the performance of state-of-the-art lip reading frameworks, which confirms the effectiveness of explicit modeling of tones for Chinese Mandarin lip reading.
\end{abstract}




\maketitle

\section{Introduction}\label{sec:introduction}
Lip reading, also known as visual speech recognition, aims to predict the sentence being spoken, given a silent video of a talking face. In noisy environments, where speech recognition is difficult, visual speech recognition offers an alternative way to understand speech. Besides, lip reading has practical potential in improved hearing aids, security, and silent dictation in public spaces. Lip reading is essentially a difficult problem, as most lip reading actuations, besides the lips and sometimes tongue and teeth, are latent and ambiguous. Several seemingly identical lip movements can produce different words.

Thanks to the recent development of deep learning, English-based lip reading methods have made great progress, at both word-level \cite{petridis2016deep, chung2016lip} and  sentence-level
\cite{assael2016lipnet, chung2017lipWild}. However, as the language of the most number of speakers, there is only a little work for Chinese Mandarin lip reading in the multimedia community. Yang et al.~\cite{yang2018lrw} present a naturally-distributed large-scale benchmark for Chinese Mandarin lip-reading in the wild, named LRW-1000, which contains 1,000 classes with 718,018 samples from more than 2,000 individual speakers. Each class corresponds to the syllables of a Mandarin word composed of one or several Chinese characters. However, they perform only word classification for Chinese Mandarin lip reading but not at the complete sentence level. LipCH-Net \cite{zhang2019understanding} is the first paper aiming for sentence-level Chinese Mandarin lip reading. LipCH-Net is a two-step end-to-end architecture, in which two deep neural network models are employed to perform the recognition of Picture-to-Pinyin (mouth motion pictures to pronunciations) and the recognition of Pinyin-to-Hanzi (pronunciations to texts) respectively. Then a joint optimization is performed to improve the overall performance.

Belong to two different language families, English and Chinese Mandarin have many differences. The most significant one might be that: Chinese Mandarin is a tone language, while English is not. The tone is the use of pitch in language to distinguish lexical or grammatical meaning - that is, to distinguish or to inflect words \footnote{\url{https://en.wikipedia.org/wiki/Tone\_(linguistics)}}. Even two words look the same on the face when pronounced, they can have different tones, thus have different meanings. For example, even though "\begin{CJK*}{UTF8}{gbsn}练习\end{CJK*}" (which means \textit{practice}) and "\begin{CJK*}{UTF8}{gbsn}联系\end{CJK*}" (which means \textit{contact}) have different meanings, but they have the same mouth movement. This increases ambiguity when lip reading. So the tone is an important factor for Chinese Mandarin lip reading.

Based on the above considerations, in this paper, we present CSSMCM, a sentence-level Chinese Mandarin lip reading network, which contains three sub-networks. Same as \cite{zhang2019understanding}, in the first sub-network, pinyin sequence is predicted from the video. Different from \cite{zhang2019understanding}, which predicts pinyin characters from video, pinyin is taken as a whole in CSSMCM, also known as syllables. As we know, Mandarin Chinese is a syllable-based language and syllables are their logical unit of pronunciation. Compared with pinyin characters, syllables are a longer linguistic unit, and can reduce the difficulty of syllable choices in the decoder by sequence-to-sequence attention-based models \cite{zhou2018syllable}. Chen et al.~\cite{chen2008seeing} find that there might be a relationship between the production of lexical tones and the visible movements of the neck, head, and mouth. Motivated by this observation, in the second sub-network, both video and pinyin sequence is used as input to predict tone. Then in the third sub-network, video, pinyin, and tone sequence work together to predict the Chinese character sequence. At last, three sub-networks are jointly finetuned to improve overall performance.

As there is no public sentence-level Chinese Mandarin lip reading dataset, we collect a new Chinese Mandarin Lip Reading dataset called CMLR based on China Network Television broadcasts containing talking faces together with subtitles of what is said.

In summary, our major contributions are as follows.
\begin{itemize}
\item We argue that tone is an important factor for Chinese Mandarin lip reading, which increases the ambiguity compared with English lip reading. Based on this, a three-stage cascade network, CSSMCM, is proposed. The tone is inferred by video and syntactic structure, and are used to predict sentence along with visual information and syntactic structure.
\item We collect a 'Chinese Mandarin Lip Reading' (CMLR) dataset, consisting of over 100,000 natural sentences from national news program "News Broadcast". The dataset will be released as a resource for training and evaluation.
\item Detailed experiments on CMLR dataset show that explicitly modeling tone when predicting Chinese sentence performs a lower character error rate.
\end{itemize}

\begin{table}[htb]
\small
\centering
\caption {Symbol Definition}
\label{table:symbol_definition}
\begin{tabular}{l | p{5.8cm}}
  \hline 
  \textbf{Symbol}     & \textbf{Definition}                                     \\ \hline     
    ${\rm GRU^v_e}$    &  GRU unit in video encoder                                \\ \hline
    ${\rm GRU^p_e}, {\rm GRU^p_d}$    &  GRU unit in pinyin encoder and pinyin decoder\\ \hline    
    ${\rm GRU^t_e}, {\rm GRU^t_d}$    &  GRU unit in tone encoder and tone decoder    \\ \hline
    ${\rm GRU^y_d}$    &  GRU unit in character decoder                            \\ \hline
    ${\rm Attention^v_p}$  & attention between pinyin decoder and video encoder. The superscript indicates the encoder and the subscript indicates the decoder.\\ \hline
    $x, y, p, t$    &  video, character, pinyin, and tone sequence              \\ \hline
    $i$         & timestep \\ \hline
    $h^v_e, h^p_e, h^t_e$ &  video encoder output, pinyin encoder output, tone encoder output                \\ \hline
    $c^v, c^p, c^t$ &  video content, pinyin content, tone content               \\ \hline
\end{tabular} 
\end{table}

\section{The Proposed Method}
In this section, we present CSSMCM, a lip reading model for Chinese Mandarin. As mention in Section \ref{sec:introduction}, pinyin and tone are both important for Chinese Mandarin lip reading. Pinyin represents how to pronounce a Chinese character and is related to mouth movement. Tone can alleviate the ambiguity of visemes (several speech sounds that look the same) to some extent and can be inferred from visible movements. Based on this, the lip reading task is defined as follow:
\begin{equation}\label{eq:1}
P(y|x) = \sum_p\sum_t P(y|p, t, x)P(t|p, x)P(p|x),
\end{equation}  
The meaning of these symbols is given in Table \ref{table:symbol_definition}.

As shown in Equation~(\ref{eq:1}), the whole problem is divided into three parts, which corresponds to pinyin prediction, tone prediction, and character prediction separately. Each part will be described in detail below.

\subsection{Pinyin Prediction Sub-network}\label{sec:pinyin_prediction_sub-network}

The pinyin prediction sub-network transforms video sequence into pinyin sequence, which corresponds to $P(p|x)$ in Equation~(\ref{eq:1}). This sub-network is based on the sequence-to-sequence architecture with attention mechanism \cite{bahdanau2015neural}. We name the encoder and decoder the video encoder and pinyin decoder, for the encoder process video sequence, and the decoder predicts pinyin sequence. The input video sequence is first fed into the VGG model \cite{chatfield2014return} to extract visual feature. The output of conv5 of VGG is appended with global average pooling \cite{lin2014network} to get the $512$-dim feature vector. Then the $512$-dim feature vector is fed into video encoder. The video encoder can be denoted as:
\begin{equation}
(h^v_e)_i = {\rm GRU}^v_e((h^v_e)_{i-1}, {\rm VGG}(x_i)).
\end{equation}
When predicting pinyin sequence, at each timestep $i$, video encoder outputs are attended to calculate a context vector $c^v_i$:
\begin{equation}
(h^p_d)_i = {\rm GRU^p_d}((h^p_d)_{i-1}, p_{i-1}),
\end{equation}
\begin{equation}
c^v_i = h^v_e \cdot {\rm Attention^v_p}((h^p_d)_i, h^v_e),
\end{equation}
\begin{equation}
P(p_i|p_{<{i}}, x) = {\rm softmax}({\rm MLP}((h^p_d)_i, c^v_i)).
\end{equation}

\subsection{Tone Prediction Sub-network}
\begin{figure}
\centering
\includegraphics[width=2.7in]{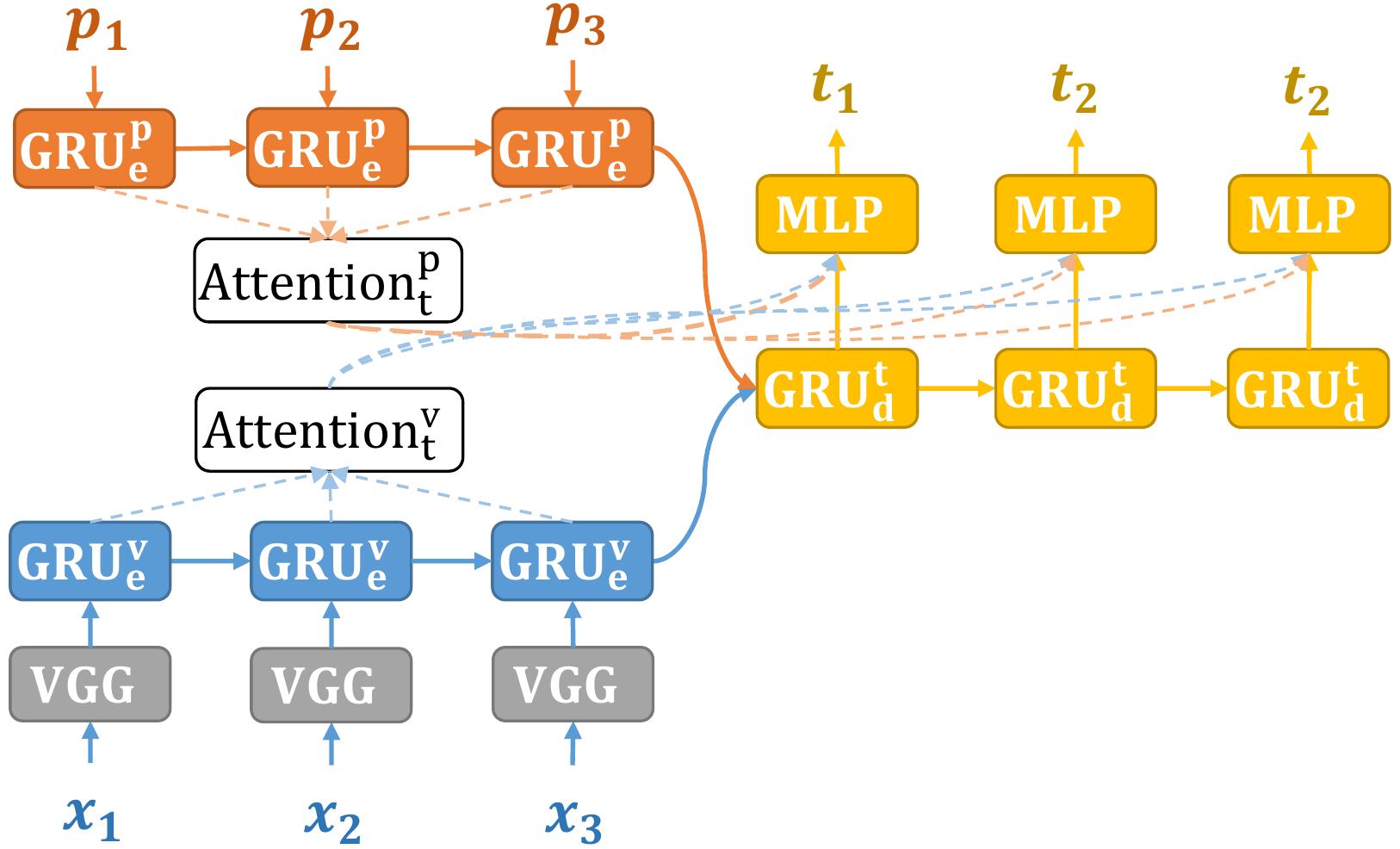}
\caption{The tone prediction sub-network.}\label{fig:network2}
\end{figure}
As shown in Equation~(\ref{eq:1}), tone prediction sub-network ($P(t|p, x)$) takes video and pinyin sequence as inputs and predict corresponding tone sequence. This problem is modeled as a sequence-to-sequence learning problem too. The corresponding model architecture is shown in Figure \ref{fig:network2}. 

In order to take both video and pinyin information into consideration when producing tone, a dual attention mechanism \cite{chung2017lipWild} is employed. Two independent attention mechanisms are used for video and pinyin sequence. Video context vectors $c^v_i$ and pinyin context vectors $c^p_i$ are fused when predicting a tone character at each decoder step.

The video encoder is the same as in Section~\ref{sec:pinyin_prediction_sub-network} and the pinyin encoder is:
\begin{equation}\label{eq:pinyin_encoder_input}
(h^p_e)_{i} = {\rm GRU^p_e}((h^p_e)_{i-1}, p_{i-1}).
\end{equation}
The tone decoder takes both video encoder outputs and pinyin encoder outputs to calculate context vector, and then predicts tones:
\begin{equation}
(h^t_d)_i = {\rm GRU^t_d}((h^t_d)_{i-1}, t_{i-1}),
\end{equation} 
\begin{equation}
c^v_i = h^v_e \cdot {\rm Attention^v_t}((h^t_d)_i, h^v_e),
\end{equation}
\begin{equation}
c^p_i = h^p_e \cdot {\rm Attention^p_t}((h^t_d)_i, h^p_e),
\end{equation}
\begin{equation}
P(t_i|t_{<i}, x, p) = {\rm softmax}({\rm MLP}((h^t_d)_i, c^v_i, c^p_i)).
\end{equation}

\subsection{Character Prediction Sub-network}
\begin{figure}
\centering
\includegraphics[width=3.2in]{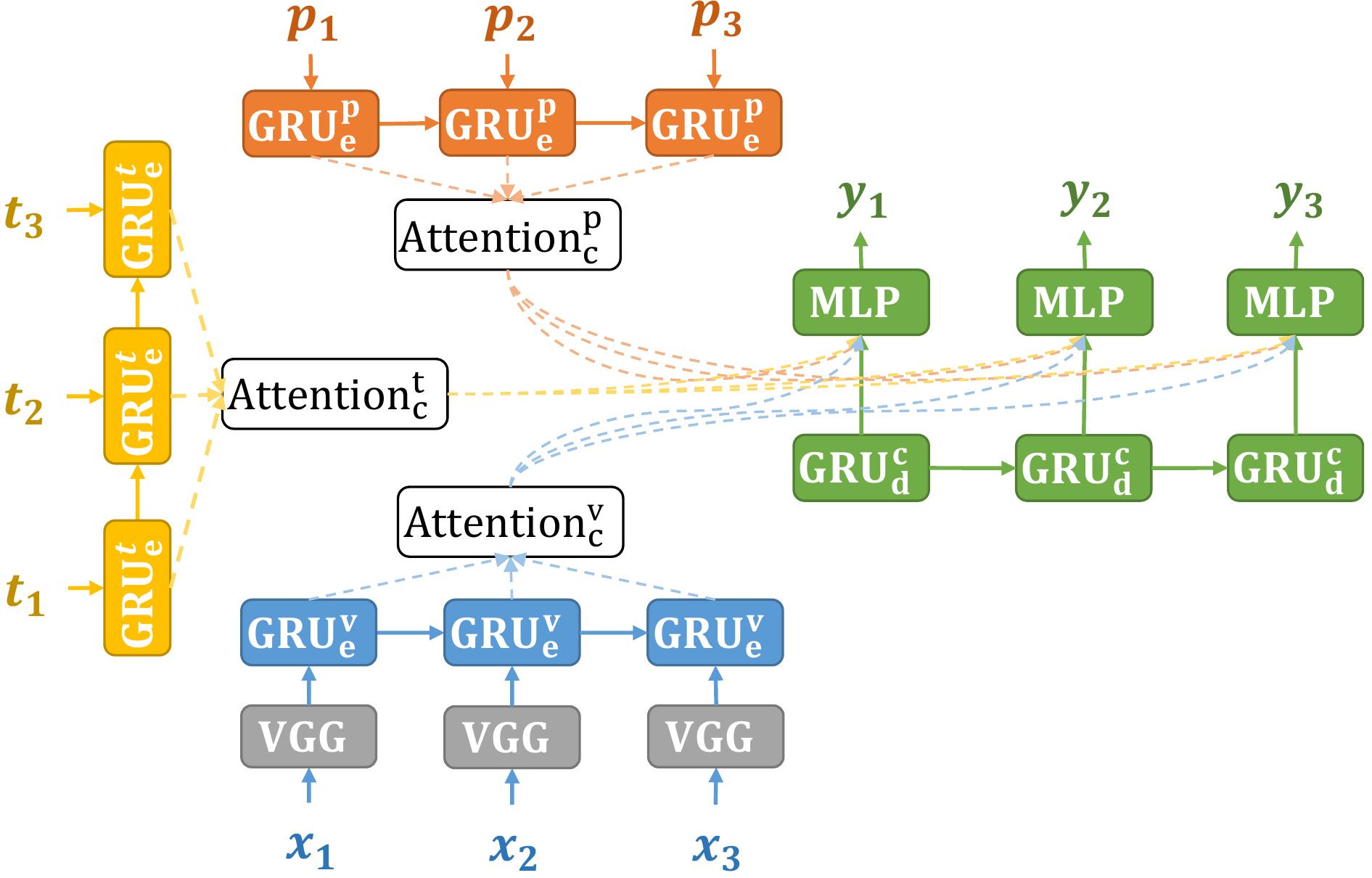}
\caption{The character prediction sub-network.}\label{fig:network3}
\end{figure}
The character prediction sub-network corresponds to $P(y|p, t, x)$ in Equation~(\ref{eq:1}).
It considers all the pinyin sequence, tone sequence and video sequence when predicting Chinese character. Similarly, we also use attention based sequence-to-sequence architecture to model this equation. Here the attention mechanism is modified into triplet attention mechanism:

\begin{equation}
(h^c_d)_i = {\rm GRU^c_d}((h^c_d)_{i-1}, y_{i-1}),
\end{equation} 
\begin{equation}
c^v_i = h^v_e \cdot {\rm Attention^v_c}((h^c_d)_i, h^v_e),
\end{equation}
\begin{equation}
c^p_i = h^p_e \cdot {\rm Attention^p_c}((h^c_d)_i, h^p_e),
\end{equation}
\begin{equation}
c^t_i = h^t_e \cdot {\rm Attention^t_c}((h^c_d)_i, h^t_e),
\end{equation}
\begin{equation}
P(c_i|c_{<i}, x, p, t) = {\rm softmax}({\rm MLP}((h^c_d)_i, c^v_i, c^p_i, c^t_i)).
\end{equation}

For the following needs, the formula of tone encoder is also listed as follows:
\begin{equation}\label{eq:tone_encoder_input}
(h^t_e)_i = {\rm GRU^t_e}((h^t_e)_{i-1}, t_{i-1}).
\end{equation}

\subsection{CSSMCM Architecture}
\begin{figure}
\centering
\includegraphics[width=3.3in]{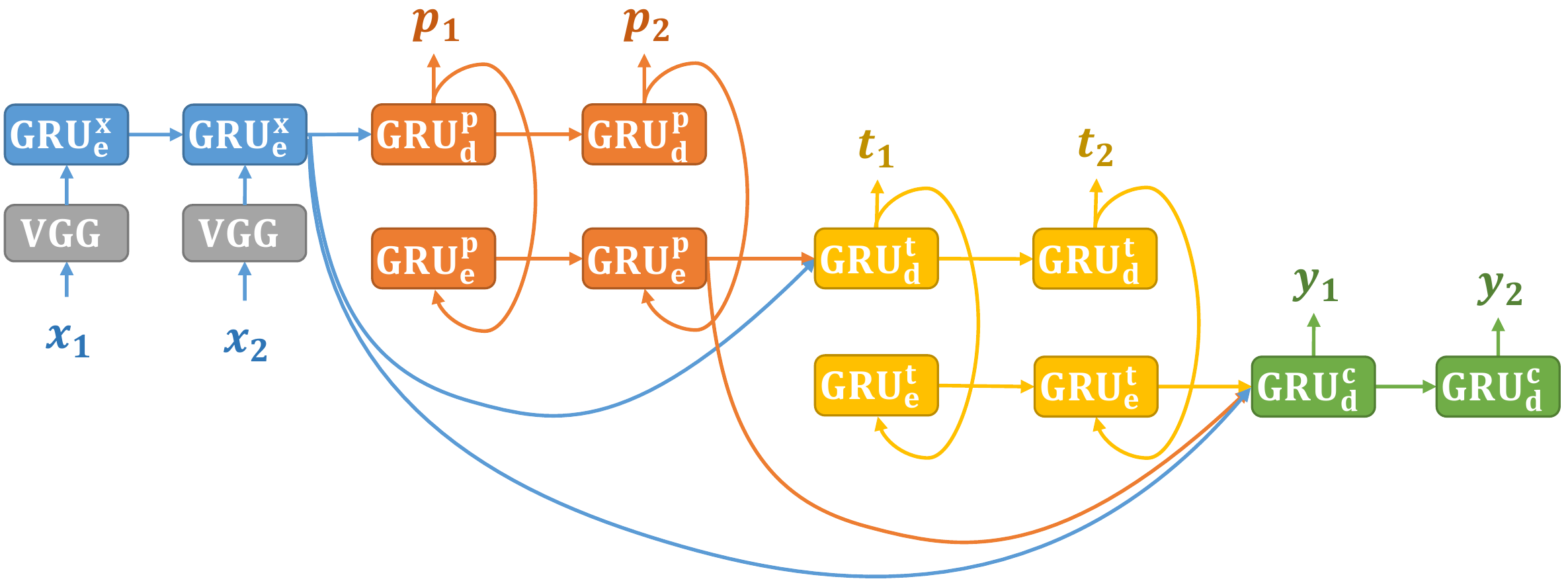}
\caption{The overall of the CSSMCM network. The attention module is omitted for sake of simplicity.}\label{fig:overall}
\end{figure}

The architecture of the proposed approach is demonstrated in Figure~\ref{fig:overall}. For better display, the three attention mechanisms are not shown in the figure. During the training of CSSMCM, the outputs of pinyin decoder are fed into pinyin encoder, the outputs of tone decoder into tone encoder:
\begin{equation}\label{eq:replaced_pinyin_encoder_input}
(h^p_e)_i = {\rm GRU^p_e}((h^p_e)_{i-1}, {\rm MLP}((h^t_d)_i, c^v_i, c^p_i)),
\end{equation}
\begin{equation}\label{eq:replaced_tone_encoder_input}
(h^t_e)_i = {\rm GRU^t_e}((h^p_e)_{i-1}, {\rm MLP}((h^c_d)_i, c^v_i, c^p_i, c^t_i)).
\end{equation}

We replace Equation~(\ref{eq:pinyin_encoder_input}) with Equation~(\ref{eq:replaced_pinyin_encoder_input}), Equation~(\ref{eq:tone_encoder_input}) with Equation (\ref{eq:replaced_tone_encoder_input}). Then, the three sub-networks are jointly trained and the overall loss function is defined as follows:
\begin{equation}
L = L_p + L_t + L_c,
\end{equation}
where $L_p, L_t$ and $L_c$ stand for loss of pinyin prediction sub-network, tone prediction sub-network and character prediction sub-network respectively, as defined below.
\begin{equation}
\begin{split}
    L_p &= - \sum_i \log P(p_i|p_{<{i}}, x),\\
    L_t &= - \sum_i \log P(t_i|t_{<i}, x, p),\\
    L_c &= - \sum_i \log P(c_i|c_{<i}, x, p, t). \\
\end{split}
\end{equation}

\subsection{Training Strategy}
To accelerate training and reduce overfitting, curriculum learning~\cite{chung2017lipWild} is employed. The sentences are grouped into subsets according to the length of less than 11, 12-17, 18-23, more than 24 Chinese characters. Scheduled sampling proposed by \cite{bengio2015scheduled} is used to eliminate the discrepancy between training and inference. At the training stage, the sampling rate from the previous output is selected from 0.7 to 1. Greedy decoder is used for fast decoding.

\section{Dataset}
In this section, a three-stage pipeline for generating the Chinese Mandarin Lip Reading (CMLR) dataset is described, which includes video pre-processing, text acquisition, and data generation. This three-stage pipeline is similar to the method mentioned in \cite{chung2017lipWild}, but considering the characteristics of our Chinese Mandarin dataset, we have optimized some steps and parts to generate a better quality lip reading dataset. The three-stage pipeline is detailed below.

\textbf{Video Pre-processing}. 
First, national news program "News Broadcast" recorded between June 2009 and June 2018 is obtained from China Network Television website. Then, the HOG-based face detection method is performed \cite{king2009dlib}, followed by an open source platform for face recognition and alignment. The video clip set of eleven different hosts who broadcast the news is captured. During the face detection step, using frame skipping can improve efficiency while ensuring the program quality.

\textbf{Text Acquisition}. 
Since there is no subtitle or text annotation in the original "News Broadcast" program, FFmpeg tools \footnote{\url{https://ffmpeg.org/}} are used to extract the corresponding audio track from the video clip set. 
Then through the iFLYTEK \footnote{\url{https://www.xfyun.cn/}} ASR, the corresponding text annotation of the video clip set is obtained. However, there is some noise in these text annotation. English letters, Arabic numerals, and rare punctuation are deleted to get a more pure Chinese Mandarin lip reading dataset.

\textbf{Data Generation}. 
The text annotation acquired in the previous step also contains timestamp information. Therefore, video clip set is intercepted according to these timestamp information, and then the corresponding word, phrase, or sentence video segment of the text annotation are obtained.
Since the text timestamp information may have a few uncertain errors, some adjustments are made to the start frame and the end frame when intercepting the video segment. It is worth noting that through experiments, we found that using OpenCV \footnote{\url{http://docs.opencv.org/2.4.13/modules/refman.html}} can capture clearer video segment than the FFmpeg tools.

Through the three-stage pipeline mentioned above, we can obtain the Chinese Mandarin Lip Reading (CMLR) dataset containing more than 100,000 sentences, 25,000 phrases, 3,500 characters. The dataset is randomly divided into training set, validation set, and test set in a ratio of 7:1:2. Details are listed in Table \ref{table:dataset}.

Further details of the dataset and the download links can be found on the web page: \url{https://www.vipazoo.cn/CMLR.html}.

\begin{table}
\caption {The CMLR dataset. Division of training, validation and test data; and the number of sentences, phrases and characters of each partition.}
\label{table:dataset}
\centering
\begin{tabular}{p{1.3cm}<{\centering} | p{1.8cm}<{\centering} | p{1.6cm}<{\centering} | p{1.8cm}<{\centering}}
  \hline \textbf{Set} & \textbf{\# sentences} & \textbf{\# phrases} & \textbf{\# characters} \\ \hline \hline 
    \textbf{Train} &  71,452  &  22,959  &  3,360    \\ \hline
    \textbf{Validation}   &  10,206  &  10,898  &  2,540    \\ \hline
    \textbf{Test}  &  20,418  &  14,478  &  2,834    \\ \hline \hline 
    
    \textbf{All}   &  102,076 &  25,633  &  3,517    \\ \hline
\end{tabular}
\end{table}

\section{Experiments}

\subsection{Implementation Details}
The input images are 64 $\times$ 128 in dimension. Lip frames are transformed into gray-scale, and the VGG network takes every 5 lip frames as an input, moving 2 frames at each timestep. For all sub-networks, a two-layer bi-direction GRU \cite{cho2014learning} with a cell size of 256 is used for the encoder and a two-layer uni-direction GRU with a cell size of 512 for the decoder. For character and pinyin vocabulary, we keep characters and pinyin that appear more than 20 times. [sos], [eos] and [pad] are also included in these three vocabularies.  The final vocabulary size is 371 for pinyin prediction sub-network, 8 for tone prediction sub-network (four tones plus a neutral tone), and 1,779 for character prediction sub-network.

The initial learning rate was 0.0001 and decreased by 50\% every time the training error did not improve for 4 epochs. CSSMCM is implemented using pytorch library and trained on a Quadro 64C P5000 with 16GB memory. The total end-to-end model was trained for around 12 days.

\subsection{Compared Methods and Evaluation Protocol}
We list here the compared methods and the evaluation protocol.

\begin{table}[htb]
\small
\centering
\caption {The detailed comparison between CSSMCM and other methods on the CMLR dataset. V, P, T, C stand for video, pinyin, tone and character. V2P stands for the transformation from video sequence to pinyin sequence. VP2T represents the input are video and pinyin sequence and the output is sequence of tone. OVERALL means to combine the sub-networks and make a joint optimization.}
\label{table:comparison}
\begin{tabular}{c | c | c | c | c}
  \hline 
  \textbf{Models} & \textbf{sub-network} & \textbf{CER} & \textbf{PER} & \textbf{TER} \\ \hline \hline
    \textbf{WAS}             &       -   &   38.93\%  &   -        &      -    \\ \hline \hline
    
    \multirow{3}{*}{\textbf{LipCH-Net-seq}}  &  V2P      &     -     &   27.96\% &     -    \\ 
                                    &  P2C      &    9.88\% &      -    &     -    \\
                                    &  OVERALL & 34.07\%    & 39.52\%   &   -    \\\hline \hline
                                    
    \multirow{4}{*}{\textbf{CSSMCM-w/o video}}&  V2P     &     -     &   27.96\% &     -    \\ 
                                    &  P2T      &    -      &      -    &     6.99\%     \\
                                    &  PT2C     &   4.70 \% &     -     &     -\\
                                    & OVERALL   & 42.23\%   & 46.67\%   &   13.14\%  \\\hline \hline
                                    
    \multirow{4}{*}{\textbf{CSSMCM}}         &   V2P     &     -     &   27.96\%  &     -    \\ 
                                    &   VP2T    &    -      &      -     &     6.14\%    \\
                                    &   VPT2C   &  3.90\%   &       -    &      - \\
                                    &   OVERALL & 32.48\%   &   36.22\%  &  10.95\%  \\\hline 
    
\end{tabular} 
\end{table}

\textbf{WAS}: The architecture used in \cite{chung2017lipWild} without the audio input. The decoder output Chinese character at each timestep. Others keep unchanged to the original implementation. 

\textbf{LipCH-Net-seq}: For a fair comparison, we use sequence-to-sequence with attention framework to replace the Connectionist temporal classification (CTC) loss \cite{graves2006connectionist} used in LipCH-Net \cite{zhang2019understanding} when converting picture to pinyin. 

\textbf{CSSMCM-w/o video}: To evaluate the necessity of video information when predicting tone, the video stream is removed when predicting tone and Chinese characters. In other word, video is only used when predicting the pinyin sequence. The tone is predicted from the pinyin sequence. Tone information and pinyin information work together to predict Chinese character.

We tried to implement the Lipnet architecture \cite{assael2016lipnet} to predict Chinese character at each timestep. However, the model did not converge. The possible reasons are due to the way CTC loss works and the difference between English and Chinese Mandarin. Compared to English, which only contains 26 characters, Chinese Mandarin contains thousands of Chinese characters. When CTC calculates loss, it first adds blank between every character in a sentence, that causes the number of the blank label is far more than any other Chinese character. Thus, when Lipnet starts training, it predicts only the blank label. After a certain epoch, "\begin{CJK*}{UTF8}{gbsn}的\end{CJK*}" character will occasionally appear until the learning rate decays to close to zero.

\begin{table*}[htp]
\small
\centering
\caption {Examples of sentences that CSSMCM correctly predicts while other methods do not. The pinyin and tone sequence corresponding to the Chinese character sentence are also displayed together. GT stands for ground truth.}
\label{table:text_generate}
\begin{tabular}{p{2.7cm}<{\centering}| c | c | c}
\hline
\textbf{Method} & \textbf{Chinese Character Sentence} & \textbf{Pinyin Sequence} & \textbf{Tone Sequence} \\ \hline \hline 

\textbf{GT}   &  \begin{CJK*}{UTF8}{gbsn}既让老百姓得实惠\end{CJK*}   &  ji rang lao bai xing de shi hui     &   4 4 3 3 4 2 2 4  \\ \hline
\textbf{WAS}            &  \begin{CJK*}{UTF8}{gbsn}介项老百姓姓事会\end{CJK*}   &  jie xiang lao bai xing xing shi hui &   4 4 3 3 4 4 4 4\\ \hline
\textbf{LipCH-Net-seq}  &  \begin{CJK*}{UTF8}{gbsn}既让老百姓的吃贵\end{CJK*}   &  ji rang lao bai xing de chi gui     &   4 4 3 3 4 0 1 4\\ \hline
\textbf{CSSMCM}          &  \begin{CJK*}{UTF8}{gbsn}既让老百姓得实惠\end{CJK*}   &  ji rang lao bai xing de shi hui     &   4 4 3 3 4 2 2 4\\ \hline \hline

\textbf{GT}   &  \begin{CJK*}{UTF8}{gbsn}有效应对当前半岛局势\end{CJK*}   &  you xiao ying dui dang qian ban dao ju shi    &   3 4 4 4 1 2 4 3 2 4    \\ \hline
\textbf{WAS}            &  \begin{CJK*}{UTF8}{gbsn}有效应对当天半岛趋势\end{CJK*}   &  you xiao ying dui dang tian ban dao qu shi    &   3 4 4 4 1 1 4 3 1 4    \\ \hline
\textbf{LipCH-Net-seq}  &  \begin{CJK*}{UTF8}{gbsn}有效应对党年半岛局势\end{CJK*}   &  you xiao ying dui dang nian ban dao ju shi    &   3 4 4 4 3 2 4 3 2 4   \\ \hline
\textbf{CSSMCM}          &  \begin{CJK*}{UTF8}{gbsn}有效应对当前半岛局势\end{CJK*}   &  you xiao ying dui dang qian ban dao ju shi    &   3 4 4 4 1 2 4 3 2 4     \\ \hline
\end{tabular}
\end{table*}

\begin{table}[htp]
\small
\centering
\caption {Failure cases of CSSMCM.}
\label{table:text_failure}
\begin{tabular}{c | c }

\hline
\multirow{2}{*}{\textbf{GT}}     & \begin{CJK*}{UTF8}{gbsn}向全球价值链中高端迈进\end{CJK*} \\
                & xiang quan qiu jia zhi lian zhong gao duan mai jin    \\ \hline
\multirow{2}{*}{\textbf{CSSMCM}} & \begin{CJK*}{UTF8}{gbsn}向全球下试联中高端迈进\end{CJK*} \\
                & xiang quan qiu xia shi lian zhong gao duan mai jin   \\ \hline \hline
  
\multirow{2}{*}{\textbf{GT}}     & \begin{CJK*}{UTF8}{gbsn}随着我国医学科技的进步\end{CJK*}  \\
                & sui zhe wo guo yi xue ke ji de jin bu   \\ 
                \hline
\multirow{2}{*}{\textbf{CSSMCM}} & \begin{CJK*}{UTF8}{gbsn}随着我国一水科技的信步\end{CJK*}  \\
                & sui zhe wo guo yi shui ke ji de jin bu   \\ 
\hline

\end{tabular}
\end{table}

For all experiments, Character Error Rate (CER) and Pinyin Error Rate (PER) are used as evaluation metrics. CER is defined as $ErrorRate = (S + D + I) / N$, where $S$ is the number of substitutions, $D$ is the number of deletions, $I$ is the number of insertions to get from the reference to the hypothesis and $N$ is the number of words in the reference. PER is calculated in the same way as CER. Tone Error Rate (TER) is also included when analyzing CSSMCM, which is calculated in the same way as above.

\subsection{Results}
Table \ref{table:comparison} shows a detailed comparison between various sub-network of different methods. Comparing P2T and VP2T, VP2T considers video information when predicting the pinyin sequence and achieves a lower error rate. This verifies the conjecture of \cite{chen2008seeing} that the generation of tones is related to the motion of the head. In terms of overall performance, CSSMCM exceeds all the other architecture on the CMLR dataset and achieves 32.48\% character error rate. It is worth noting that CSSMCM-w/o video achieves the worst result (42.23\% CER) even though its sub-networks perform well when trained separately. This may be due to the lack of visual information to support, and the accumulation of errors. CSSMCM using tone information performs better compared to LipCH-Net-seq, which does not use tone information. The comparison results show that tone is important when lip reading, and when predicting tone, visual information should be considered.

Table \ref{table:text_generate} shows some generated sentences from different methods. CSSMCM-w/o video architecture is not included due to its relatively lower performance. These are sentences other methods fail to predict but CSSMCM succeeds. The phrase "\begin{CJK*}{UTF8}{gbsn}实惠\end{CJK*}" (which means \textit{affordable}) in the first example sentence, has a tone of 2, 4 and its corresponding pinyin are \textit{shi, hui}. WAS predicts it as "\begin{CJK*}{UTF8}{gbsn}事会\end{CJK*}" (which means \textit{opportunity}). Although the pinyin prediction is correct, the tone is wrong. LipCH-Net-seq predicts "\begin{CJK*}{UTF8}{gbsn}实惠\end{CJK*}" as "\begin{CJK*}{UTF8}{gbsn}吃贵\end{CJK*}" (not a word), which have the same finals "\textit{ui}" and the corresponding mouth shapes are the same. It's the same in the second example. "\begin{CJK*}{UTF8}{gbsn}前, 天, 年\end{CJK*}" have the same finals and mouth shapes, but the tone is different. 

These show that when predicting characters with the same lip shape but different tones, other methods are often unable to predict correctly. However, CSSMCM can leverage the tone information to predict successfully.

Apart from the above results, Table \ref{table:text_failure} also lists some failure cases of CSSMCM. The characters that CSSMCM predicts wrong are usually homophones or characters with the same final as the ground truth. In the first example, "\begin{CJK*}{UTF8}{gbsn}价\end{CJK*}" and "\begin{CJK*}{UTF8}{gbsn}下\end{CJK*}" have the same final, \textit{ia}, while "\begin{CJK*}{UTF8}{gbsn}一\end{CJK*}" and "\begin{CJK*}{UTF8}{gbsn}医\end{CJK*}" are homophones in the second example. Unlike English, if one character in an English word is predicted wrong, the understanding of the transcriptions has little effect. However, if there is a character predicted wrong in Chinese words, it will greatly affect the understandability of transcriptions. In the second example, CSSMCM mispredicts "\begin{CJK*}{UTF8}{gbsn}医学\end{CJK*}" ( which means \textit{medical}) to "\begin{CJK*}{UTF8}{gbsn}一水\end{CJK*}" (which means \textit{all}). Although their first characters are pronounced the same, the meaning of the sentence changed from \textit{Now with the progress of medical science and technology in our country} to \textit{It is now with the footsteps of China's Yishui Technology}.

\subsection{Attention Visualisation}
\begin{figure*}[htbp]
\centering
\subfigure[]{
\begin{minipage}[t]{0.5\linewidth}
\centering
\includegraphics[width=3.0in]{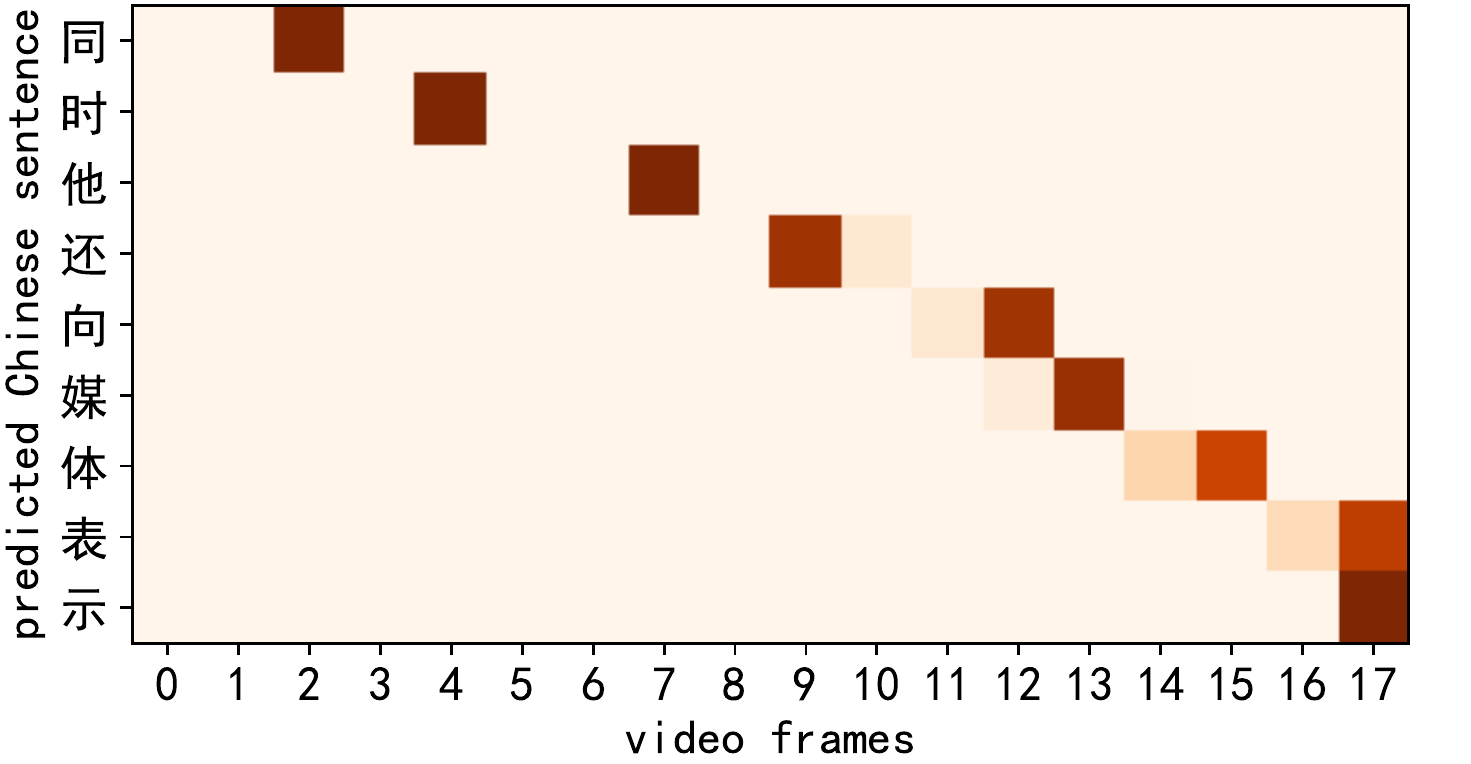}
\end{minipage}%
}%
\subfigure[]{
\begin{minipage}[t]{0.5\linewidth}
\centering
\includegraphics[width=3.0in]{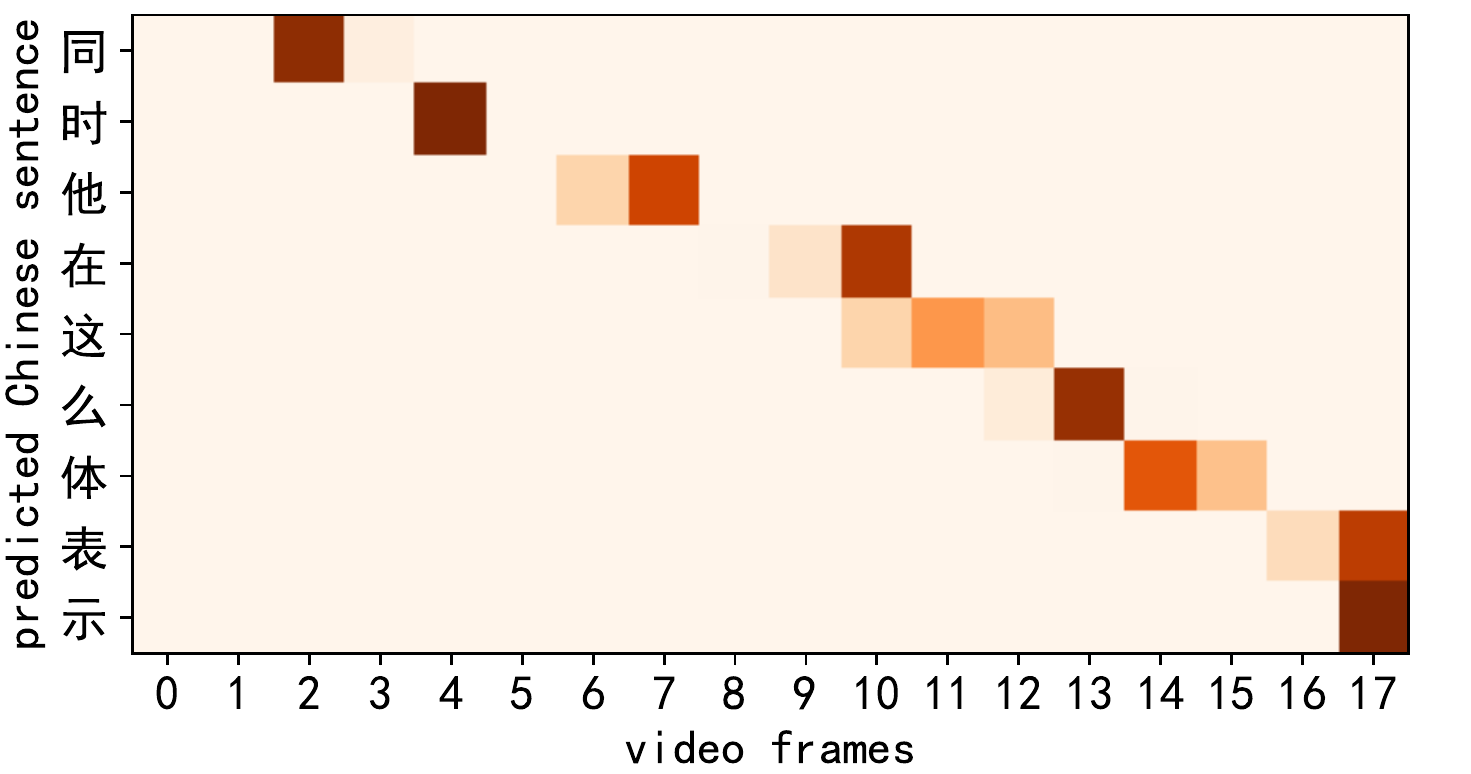}
\end{minipage}%
}%
\centering
\caption{Video-to-text alignment using CSSMCM (a) and WAS (b).}
\label{fig:video_text_alignment}
\end{figure*}

\begin{figure}[htp]
\centering
\includegraphics[width=2.3in]{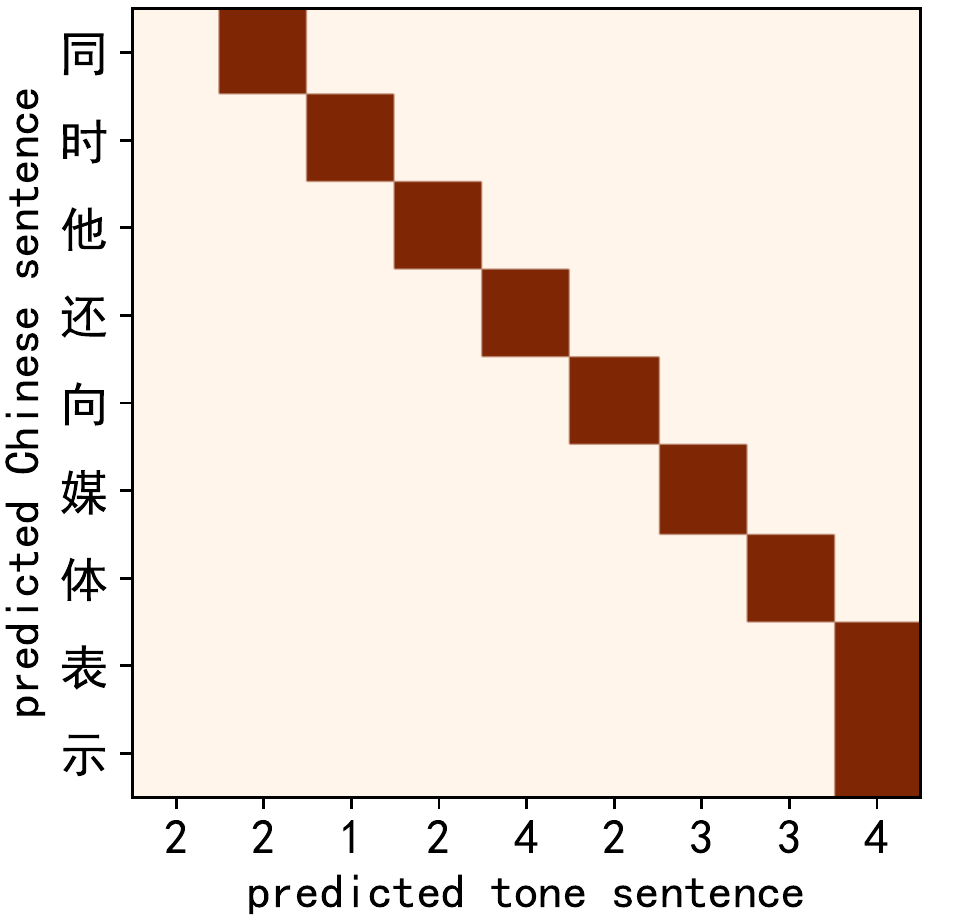}
\caption{Aligenment between output characters and predicted tone sequences using CSSMCM.}
\label{fig:CSSMCM_tone}
\end{figure}


Figure \ref{fig:video_text_alignment} (a) and Figure \ref{fig:video_text_alignment} (b) visualise the alignment of video frames and Chinese characters predicted by CSSMCM and WAS respectively. The ground truth sequence is "\begin{CJK*}{UTF8}{gbsn}同时他还向媒体表示\end{CJK*}". Comparing Figure \ref{fig:video_text_alignment} (a) with Figure \ref{fig:video_text_alignment} (b), the diagonal trend of the video attention map got by CSSMCM is more obvious. The video attention is more focused where WAS predicts wrong, i.e. \ the area corresponding to "\begin{CJK*}{UTF8}{gbsn}还向\end{CJK*}". Although WAS mistakenly predicts the "\begin{CJK*}{UTF8}{gbsn}媒体\end{CJK*}" as "\begin{CJK*}{UTF8}{gbsn}么体\end{CJK*}", the "\begin{CJK*}{UTF8}{gbsn}媒体\end{CJK*}" and the "\begin{CJK*}{UTF8}{gbsn}么体\end{CJK*}" have the same mouth shape, so the attention concentrates on the correct frame.

It's interesting to mention that in Figure \ref{fig:CSSMCM_tone}, when predicting the $i$-th character, attention is concentrated on the $i+1$-th tone. This may be because attention is applied to the outputs of the encoder, which actually includes all the information from the previous $i+1$ timesteps. The attention to the tone of $i+1$-th timestep serves as the language model, which reduces the options for generating the character at $i$-th timestep, making prediction more accurate.

\section{Summary and Extension}
In this paper, we propose the CSSMCM, a Cascade Sequence-to-Sequence Model for Chinese Mandarin lip reading. CSSMCM is designed to predicting pinyin sequence, tone sequence, and Chinese character sequence one by one. When predicting tone sequence, a dual attention mechanism is used to consider video sequence and pinyin sequence at the same time. When predicting the Chinese character sequence, a triplet attention mechanism is proposed to take all the video sequence, pinyin sequence, and tone sequence information into consideration. CSSMCM consistently outperforms other lip reading architectures on the proposed CMLR dataset. 

Lip reading and speech recognition are very similar. In Chinese Mandarin speech recognition, there have been kinds of different acoustic representations like syllable initial/final approach, syllable initial/final with tone approach, syllable approach, syllable with tone approach, preme/toneme approach \cite{chen1997new} and Chinese Character approach \cite{zhou2018a}. In this paper, the Chinese character is chosen as the output unit. However, we find that the wrongly predicted characters severely affect the understandability of transcriptions. Using larger output units, like Chinese words, maybe can alleviate this problem. 

\section{acknowledgements}
This work is supported by  National Key Research and Development Program (2018AAA0101503) , National Natural Science Foundation of China (61572428,U1509206),  Key Research and Development Program of Zhejiang Province (2018C01004), and the Program of International Science and Technology Cooperation (2013DFG12840).

\newpage
\bibliographystyle{ACM-Reference-Format}
\bibliography{acmart}

\end{document}